\documentclass{article}
\usepackage{amsmath,graphicx,mlspconf,enumitem,amssymb,tikz}

\usepackage{color,soul}

\copyrightnotice{\fbox{\parbox{0.99\textwidth}{{\copyright}2024 IEEE. Personal use of this material is permitted. Permission from IEEE must be obtained for all other uses, in any current or future media, including reprinting/republishing this material for advertising or promotional purposes, creating new collective works, for resale or redistribution to servers or lists, or reuse of any copyrighted component of this work in other works.}}}
\let\OLDthebibliography\thebibliography
\renewcommand\thebibliography[1]{
  \OLDthebibliography{#1}
  \setlength{\parskip}{0pt}
  \setlength{\itemsep}{0pt plus 0.1ex}
}

\toappear{2024 IEEE International Workshop on Machine Learning for Signal Processing, Sept.\ 22--25, 2024, London, UK}

\def\x{{\mathbf x}}

\def\X{{\mathbf X}}

\title{Implicit Neural Representations for Simultaneous Reduction and Continuous Reconstruction of Multi-Altitude Climate Data}
\name{%
\begin{tabular}{@{}c@{}}
Alif Bin Abdul Qayyum$^{\star}$\thanks{This work was supported in part by the Department of Energy (DOE) award DE-SC0012704.} \qquad 
Xihaier Luo$^{\dagger}$\\ 
Nathan M. Urban$^{\dagger}$ \qquad 
Xiaoning Qian$^{\star \dagger}$ \qquad 
Byung-Jun Yoon$^{\star \dagger}$
\end{tabular}}
\address{
$^{\star}$ Department of Electrical and Computer Engineering, Texas A\&M University%,\\ College Station, TX
\\
$^{\dagger}$Computational Science Initiative, Brookhaven National Laboratory%, Upton, NY\\
}

\begin{document}
% \ninept

\maketitle
% \copyrightnotice
% \vspace{-6mm}
\begin{abstract}
\vspace{-1.5mm}
The world is moving towards clean and renewable energy sources, such as wind energy, in an attempt to reduce greenhouse gas emissions that contribute to global warming. To enhance the analysis and storage of wind data, we introduce a deep learning framework designed to simultaneously enable effective dimensionality reduction and continuous representation of multi-altitude wind data from discrete observations. The framework consists of three key components: dimensionality reduction, cross-modal prediction, and super-resolution. We aim to: (1) improve data resolution across diverse climatic conditions to recover high-resolution details; (2) reduce data dimensionality for more efficient storage of large climate datasets; and (3) enable cross-prediction between wind data measured at different heights. Comprehensive testing confirms that our approach surpasses existing methods in both super-resolution quality and compression efficiency.
\end{abstract}
\vspace{-2mm}
\vspace{-4.1mm}
\section{Introduction}
\label{sec:intro}
\vspace{-3mm}
% These guidelines include complete descriptions of the fonts, spacing, and
% related information for producing your proceedings manuscripts. Please follow
% them. 
% Papers should not be longer than \textbf{6 pages}, including all text, figures, and references.

% The copyright notice is required for camera-ready paper submission. We will update it after IEEE send us the notice on website. The initial paper submission does not require copyright notice information.

% The escalating challenge of climate change necessitates immediate and strategic action to mitigate its impacts and steer towards sustainable development~\cite{rolnick2022tackling,kaack2022aligning}. 
As the earth system faces increasing temperatures, rising sea levels, and extreme weather events, the shift toward renewable energy sources emerges as a crucial countermeasure. Among these, wind energy stands out for its potential to provide a clean, inexhaustible power supply that significantly reduces greenhouse gas emissions. However, the deployment and optimization of wind energy encounter a variety of challenges. %\newline

\textit{Challenge 1: Resolution Inadequacy}. Identifying the most suitable sites for wind turbines necessitates data with a resolution as detailed as 1 square kilometer or finer~\cite{irrgang2021towards,kashinath2021physics}. Yet, the resolution offered by most wind farm simulations and numerical models falls short of this requirement, thus hampering our ability to make decisions that optimize the efficiency and output of wind energy initiatives. %\newline

\textit{Challenge 2: Excessive Data Volume}. It should be noted substantial memory and hardware requirements are needed to store and process the voluminous data generated from field measurements and enhanced simulations~\cite{klower2021compressing,huang2023compressing}. %\newline

\textit{Challenge 3: Cross-Modal Inference}. Establishing wind measurement stations in specific areas can pose challenges due to the high expenses associated with transportation and maintenance, necessitating cross-modal inference, such as estimating wind speed at higher altitudes based on wind speed readings closer to the ground. %\newline

Fortunately, advancements in deep learning offer promising avenues to overcome these hurdles. Deep super-resolution techniques can enhance low-resolution data, providing the detailed representations needed for precise analysis~\cite{gao2022earthformer,diaconu2022understanding}. Concurrently, deep learning-based data reduction compresses extensive datasets into latent formats, easing memory and hardware demands. Nonetheless, most existing deep learning approaches are grid-based and fall short of offering a continuous representation of wind fields~\cite{nguyen2023climax,requena2021earthnet2021}. Since wind fields are inherently continuous, there's a critical need for methodologies that can generate and work with continuous data representations \cite{luo2023reinstating,luo2024continuous}. As a result, to speed up the utilization of wind energy, there is a pressing need to estimate continuous wind pattern from reduced low  dimensional, discontinuous data; and also achieve this in a cross modal fashion where we can estimate wind pattern at inaccessible or expensive spaces from available data at accessible spaces. %\newline

This study introduces a novel deep learning model tailored for efficient wind data reduction and reconstruction through super-resolution with implicit neural network, addressing challenges in climatological analysis for wind energy optimization in a multi-modal fashion. Due to the combined reduction and super-resolution aspects, our super-resolution task is more challenging in a practical sense.
Overall, our contributions are as follows:
% \vspace{-1mm}
\begin{itemize}[leftmargin=0.9em, itemindent=0em, noitemsep, nolistsep]
\item We introduce GEI-LIIF, an innovative super-resolution strategy that leverages implicit neural networks, along with global encoding to improve local methods, enabling the learning of continuous, high-resolution data from its discrete counterparts.
% \item We develop various self- and cross-encoders to convert high-resolution data into a lower resolution form. This deep learning approach reduces dimensions effectively, leveraging information within and between different modes, using these encoders.
\item We propose a novel latent loss function to learn modality specific low-dimensional representations irrespective of the input modality, for cross-modal representation learning. Unlike unified latent representation learning \cite{clue}, our approach bypasses the need for a modality classifier model by learning modality specific low dimensional representation.
\end{itemize}
\vspace{-4mm}
\section{Related Work}
\label{sec:related_work}
\vspace{-3.5mm}
We briefly examine current studies across three pivotal areas: climate downscaling, super-resolution, and implicit neural representation.

% % \vspace{-3.5mm}
% \subsection{Climate Downscaling}
% % \vspace{-2mm}
% Climate downscaling, allows the translation of global climate model outputs into finer, local-scale projections \cite{keller2022downscaling,harder2023hard}. This is essential for assessing impacts of climate change on specific regions. Dynamical downscaling employs high-resolution regional climate models to simulate local climate, relying on atmospheric physics for detailed projections, including wind, temperature, and precipitation. On the contrary, statistical downscaling uses historical data to establish empirical links between large-scale atmospheric conditions and local climate variables. Dynamic downscaling, being comprehensive, demands significant computational power and relies heavily on accuracy of global climate model data \cite{chau2021deconditional, chen2022rainnet}, whereas the later is more computationally efficient and faster but assumes historical relationships will persist, disregarding changes in climate variability and extremes \cite{groenke2020climalign,liu2020climate}. 
\textit{Climate Downscaling}, an essential task for assessing impacts of climate change on specific regions, allows the translation of global climate model outputs into finer, local-scale projections, either through dynamic simulation of local climate from high-resolution regional climate model, or through establishing empirical links between large-scale atmospheric conditions and local climate variables using historical data \cite{keller2022downscaling, harder2023hard}. Different variations of these traditional approaches face issues, either accurate modeling at the expense of significant computational power and heavy reliance on accurate global climate data, or computational efficiency at the expense of failing to capture climate variability and extremes \cite{groenke2020climalign, liu2020climate}. 

% % \vspace{-4mm}
% \subsection{Deep Learning-based Super-Resolution}
% % \vspace{-2mm}
\textit{Super-Resolution}, on the other hand, through deep learning is revolutionizing the enhancement of climate data, both in spatial and time domain, delivering unparalleled detail and precision \cite{gao2022earthformer,diaconu2022understanding}. 
% Convolutional neural networks are utilized for refining spatial data, while generative adversarial networks excel in crafting realistic, high-definition images~\cite{vandal2017deepsd,requena2021earthnet2021,stengel2020adversarial}. Recurrent neural networks, including long short-term memory units and gated recurrent units, are employed for temporal data enhancement in climate science~\cite{gao2022earthformer,diaconu2022understanding}. 
In spite of specializing in analysis of complex patterns in climate data through enhancing model accuracy and details beyond traditional downscaling, these methods often rely on fixed resolutions, highlighting the need for models that offer resolution-independent, continuous climate pattern representations~\cite{luo2023reinstating}. 
% % \vspace{-6mm}
% \subsection{Implicit Neural Representation}
% % \vspace{-2mm}

\textit{Implicit Neural Representation (INR)} uses neural networks to model continuous signals, transcending traditional discrete methods like pixel and voxel grids~\cite{xie2022neural,huang2023compressing}. This approach has recently made significant strides in climate data analysis, enabling high-resolution reconstructions beyond fixed enhancement scales \cite{luo2024continuous, schwarz2023modality}. 
% In~\cite{luo2024continuous}, a context-aware indexing mechanism was introduced to enhance the efficiency of INR in reconstructing fields from sparse observations. In~\cite{schwarz2023modality}, a novel compression algorithm is introduced, utilizing INR within a universal approach to data handling. This method, fusing latent encoding with principles of sparsity, effectively generates compact yet comprehensive latent depictions of ERA5 climate data. 
In spite of having capabilities in precise, scalable data representation, traditional INRs focus only on continuous representation of single modal data, such as wind speed data measured at a specific height, necessitating continuous representation learning in cross-modal scenarios. 
% % \vspace{-2mm}
% \subsection{Multimodal Learning}
% % \vspace{-2mm}
% Deep multi-modal learning transforms machine perception by concurrently integrating diverse data sources such as text, images, audio, and video, unlike of traditional single-modal approaches. Multi-modal deep learning has shown impressive performance across different domains~\cite{HurricaneForecasting, glue, clue, mmdiffusion}. 
% GLUE~\cite{glue} proposes a combination of variational autoencoder (VAE) and a knowledge-based guidance graph to model the interaction across multiple modalities. MM-Diffusion~\cite{mmdiffusion} proposes a novel U-Net architecture capable of joint denoising process across multiple modalities simultaneously with a random-shift based attention block for accurate cross-modal alignment. CLUE~\cite{clue} proposes `cross-encoders' to learn cross-modal representation along with single-modal representation. 
% Advancements in deep multi-modal learning suggests its capability to model multi-modal climate data, such as predicting wind speed at one altitude based on data from a different altitude.
\vspace{-6mm}
\section{Method}
\label{sec:methodology}
\vspace{-2mm}
\begin{figure*}[tbp]
\begin{center}
    \includegraphics[width=0.75\textwidth]{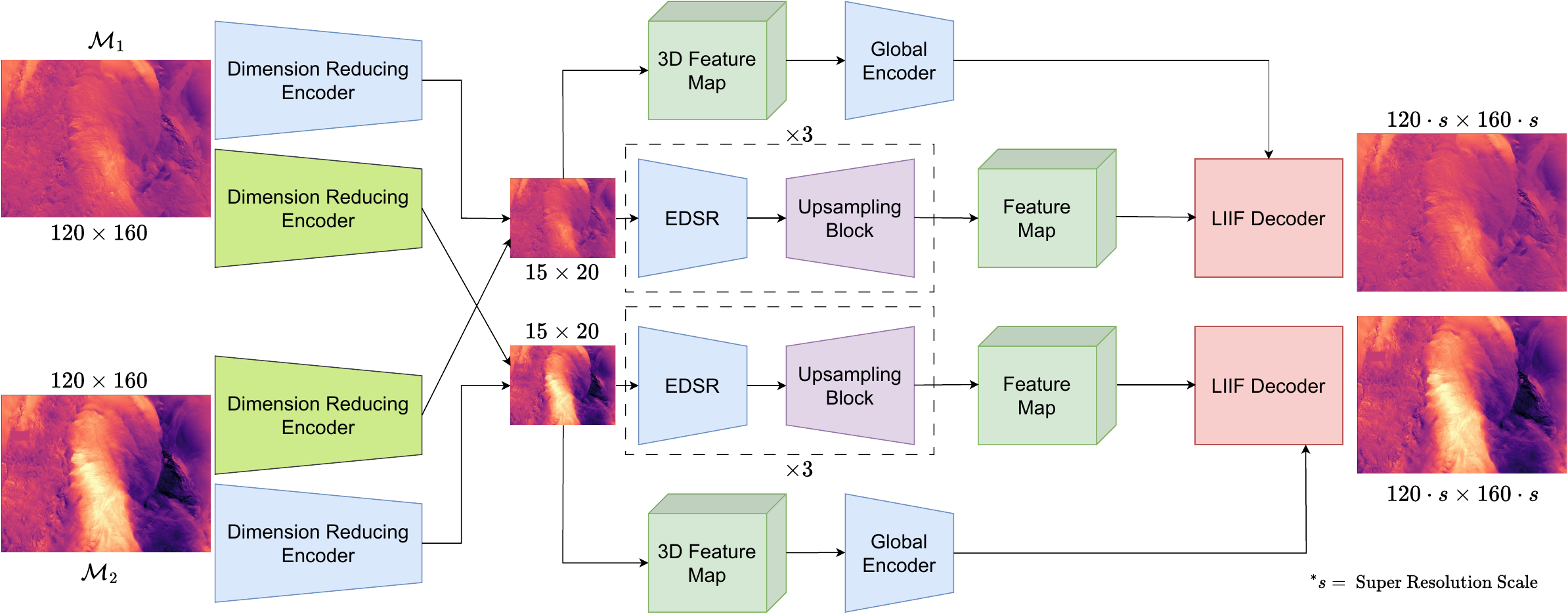}
    \vspace{-2.5mm}
    \caption{Illustration of the proposed methodology for joint dimensionality reduction and super-resolution.}
    \label{fig:methodology}
\end{center}
\vspace{-6mm}
\end{figure*}
\subsection{Problem Statement}
\vspace{-2mm}
Let, $\X$ be a data instance of a multi-modal dataset with $K$ different modalities, $\X = \{\X_1, \X_2, \cdots, \X_{K}\}$. 
% Our goal is to achieve data reduction and continuous super-resolution of this data instance, both in intra- and inter-modal scenario. Intra-modal scenario refers to the case where data reduction and continuous super-resolution from reduced dimensional data is achieved in the same modality, whereas inter-modal scenario refers to the case where they are done in different modalities.
% % % \vspace{-2mm}
% \begin{itemize}
%     \item Intra-modal scenario indicates the case where our goal is to achieve data reduction and continuous superresolution from reduced data dimension both in the same modality.
%     \item Inter-modal scenario indicates the case where our goal is to achieve data reduction in one modality and continuous superresolution from reduced data in a different modality.
% \end{itemize} 
% % % \vspace{-1mm}
% % \textcolor{blue}{@Alif please define these two scenarios Maybe it is better for you move 3.2.1 to here. RESPONSE: I believe the problem statement should not be specific to two modalities. That's why I would like to keep those texts there.}
Let, $\X_{k}^H $ be a discrete high resolution representation of this data instance with dimension $c_H \times h_H \times w_H$ in modality $k$, where $c_H$, $h_H$ and $w_H$ denote channel depth, height and width of the discrete high resolution data dimension. Now the targets to be achieved are listed below:
% % \vspace{-1mm}
\begin{itemize}[leftmargin=0.9em, itemindent=0em, noitemsep, nolistsep]
    \item \textbf{Dimension Reduction}: Extract $\X_{k}^L$, a discrete low resolution representation of $\X_{k}^H$, with dimension $c_L \times h_L \times w_L$, where $c_L$, $h_L$ and $w_L$ denote channel depth, height and width of the discrete low resolution data dimension. With dimension reduction factor $d$, $h_L = \dfrac{h_H}{d}, w_L = \dfrac{w_H}{d}$.
    \item \textbf{Continuous Representation}:  Extract a continuous super-resolution representation $\X_k^{S}$ from its corresponding discrete low representation $\X_{k}^L$, with dimension $c_{H} \times h_{S} \times w_{S}$, where $h_{S}$ and $w_{S}$ denote height and width of the continuous super-resolution data dimension. If the the super-resolution scale is $s \in \mathbb{R}$, then $h_{S} = s \cdot h_H, w_{S} = s \cdot w_H$. 
    % Extraction of continuous super-resolution from a reduced data dimension makes the task even more challenging compared to traditional super-resolution tasks. 
    \item \textbf{Modality Transfer}: Extract $\X_l^{S}$ from $\X_k^H$, where $k\neq l$. 
    % That means we need to extract the continuous super-resolution data to one modality from the discrete low resolution representation from another modality.
\end{itemize}
Super-resolution from a reduced data dimension across different modalities makes the task even more challenging compared to traditional super-resolution tasks. 
% \begin{figure}
% \begin{center}
%     \includegraphics[width=0.45\textwidth]{figs/Method-Image.pdf}
%     % % \vspace{-4mm}
%     \caption{Illustration of the problem statement.}
%     \label{fig:problem-statement}
% \end{center}
% % % \vspace{-8mm}
% \end{figure}
\vspace{-2mm}
\subsection{Model Architecture}
\vspace{-2mm}
The model architecture constitutes (i) dimension reducing encoders to encode high resolution weather data into a low resolution space, (ii) feature encoders that learn the spatial features from the low resolution representations, and (iii) implicit neural network decoders that use the extracted features by the feature encoders and predicts the wind data for that specific coordinate. We worked with 2 different modalities, $\mathcal{M}_1$ and $\mathcal{M}_2$. Figure \ref{fig:methodology} summarizes the proposed methodology. We define each modality as wind data at different heights from the ground. Specifically, weather data at $h_1$ units above from the ground is considered as modality $\mathcal{M}_1$, similarly $\mathcal{M}_2$ constitutes data at $h_2$ units above the ground, conditioned on $h_1 \neq h_2$. 
% A detailed description is provided in Section \ref{subsubsec:dataset}.
\vspace{-2mm}
\subsubsection{Dimension Reducing Encoder}
\label{subsec:dim-red-enc}
\vspace{-2mm}
Two different kinds of convolutional neural network based dimension reduction encoders, self-encoders and cross-modal encoders, with the similar architecture, encodes high resolution data to low resolution space. Self encoders convert the high resolution data from one modality to its corresponding low resolution representation, whereas the cross-modal encoders converts it into a different modality. Let $\mathcal{M}_1^H \in \mathbb{R}^{c_H \times h_H \times w_H}$ and $\mathcal{M}_2^H \in \mathbb{R}^{c_H \times h_H \times w_H}$ be the high resolution data space of $\mathcal{M}_1$ and $\mathcal{M}_2$ correspondingly. Similarly $\mathcal{M}_1^L \in \mathbb{R}^{c_L \times h_L \times w_L}$ and $\mathcal{M}_2^L \in \mathbb{R}^{c_L \times h_L \times w_L}$ be the low resolution data space of $\mathcal{M}_1$ and $\mathcal{M}_2$ correspondingly. Then, self encoders can be defined as
% \vspace{-3mm}
\begin{equation}
\mathbf{E}_1^{1}:\mathcal{M}_1^H \to \mathcal{M}_1^L \text{ and } \mathbf{E}_2^{2}:\mathcal{M}_2^H \to \mathcal{M}_2^L % \vspace{-2mm}
\end{equation}
% % \vspace{-1mm}
and the cross-modal encoders can be defined as
% \vspace{-3mm}
\begin{equation}
\mathbf{E}_1^{2}:\mathcal{M}_1^H \to \mathcal{M}_2^L\text{ and } \mathbf{E}_2^{1}:\mathcal{M}_2^H \to \mathcal{M}_1^L % \vspace{-2mm}
\end{equation}
% % \vspace{-1mm}
The architecture of the encoder is inspired from the architecture proposed in the downsampling part of the invertible UNet~\cite{iunets}.
\vspace{-2mm}
\subsubsection{Local Implicit Image Function based Decoder}
\label{subsec:liif-dec}
\vspace{-2mm}
Local implicit image function (LIIF) based decoder is a coordinate based decoding approach which takes the coordinate and the deep features around that coordinate as inputs and outputs the value for that corresponding coordinate \cite{liif}. Due to the continuous nature of spatial coordinates, LIIF-based decoder can decode into arbitrary resolution. 
% \vspace{-2mm}
\begin{equation}
\begin{split}
    &\mathbf{FE}_1:\mathcal{M}_1^L \in \mathbb{R}^{c_L \times h_L \times w_L} \to \mathcal{M}_1^F \in \mathbb{R}^{c_F \times h_F \times w_F} \\
    & \mathbf{FE}_2:\mathcal{M}_2^L \in \mathbb{R}^{c_L \times h_L \times w_L} \to \mathcal{M}_2^F \in \mathbb{R}^{c_F \times h_F \times w_F} 
\end{split} 
\end{equation}
% % \vspace{-2mm}
Here, $\mathbf{FE}_1$ and $\mathbf{FE}_2$ are two EDSR-based \cite{edsr} feature encoders for modalities $\mathcal{M}_1$ and $\mathcal{M}_2$ into the encoded feature space $\mathcal{M}_1^F$ and $\mathcal{M}_2^F$ respectively. We use $c_F$, $h_F$, $w_F$ to denote channel depth, height and width of the corresponding encoded feature space. Let $\mathbf{X}_c$ be a 2-D coordinate space. We follow the feature extraction method discussed in LIIF \cite{liif}. Let the LIIF based feature extractor be
% \vspace{-2mm}
\begin{equation}
\mathbf{F}_{ex}:\mathcal{M}_k^F \times \X_c \to \mathcal{LF}_k^F \in \mathbb{R}^p % \vspace{-1mm}
\end{equation}
% % \vspace{-1mm}
where $p$ is the extracted feature dimension, and extracted feature at coordinate $\x_c \in \X_c$ is $f^{\x_c}_k$. Decoders are functions that take the encoded features, $f^{\x_c}_k$ at specific coordinate, $\x_c$ as input. For example,
% \vspace{-2mm}
\begin{equation}
\begin{split}
    & \mathbf{D}_1:\mathcal{LF}_1^F \in \mathbb{R}^p \to \mathbf{X}^C \in \mathbb{R} \\
    & \mathbf{D}_2:\mathcal{LF}_2^F \in \mathbb{R}^p \to \mathbf{X}^C \in \mathbb{R} % \vspace{-4mm}
\end{split}
\end{equation}
% % \vspace{-2mm}
are two coordinate based decoders for modalities $\mathcal{M}_1$ and $\mathcal{M}_2$.
% $\mathbf{D}_1:\mathbf{x} \in \mathbb{R}^2, \mathcal{M}_1^F \in \mathbb{R}^{c_F \times h_F \times w_F} \to \mathbf{X}^C \in \mathbb{R}$, $\mathbf{D}_1:\mathbf{x} \in \mathbb{R}^2, \mathcal{M}_2^F \in \mathbb{R}^{c_F \times h_F \times w_F} \to \mathbf{X}^C \in \mathbb{R}$ are two coordinate based decoders for modalities $\mathcal{M}_1$ and $\mathcal{M}_2$.
\vspace{-2mm}
\subsubsection{Global Encoding Incorporated LIIF}
\label{subsec:global-pos-encoder}
\vspace{-2mm}
Global encoders are functions of the low-resolution representation. $\mathbf{G}_1:\mathcal{M}_1^L \in \mathbb{R}^{c_L \times h_L \times w_L} \to \mathcal{GF}_1^F \in \mathbb{R}^g$ and $\mathbf{G}_2:\mathcal{M}_2^L \in \mathbb{R}^{c_L \times h_L \times w_L} \to \mathcal{GF}_2^F \in \mathbb{R}^g$ are global encoders for modalities $\mathcal{M}_1$ and $\mathcal{M}_2$, with $g$ as the dimension of the global encoding. Unlike the local implicit neural network based decoder proposed in LIIF \cite{liif}, our proposed GEI-LIIF(Global Encoding Incorporated Local Implicit Image Function) based modality specific decoder, $\mathbf{D}_k$ is a function of two features:
\begin{itemize}[leftmargin=0.9em, itemindent=0em, noitemsep, nolistsep]
    \item Extracted local feature at coordinate $\x_c \in \X_c$ through
    % \vspace{-1mm}
    \begin{equation}
    \mathbf{F}_{ex}:\mathcal{M}_k^F \times \X_c \to \mathcal{LF}_k^F \in \mathbb{R}^d, f^{\x_c}_k
    % % \vspace{-1mm}
    \end{equation}
    \item Extracted global feature $gf_k$ through 
    % \vspace{-1mm}
    \begin{equation}
    \mathbf{G}_k:\mathcal{M}_k^L \in \mathbb{R}^{c_L \times h_L \times w_L} \to \mathcal{GF}_k^F \in \mathbb{R}^g
    % % \vspace{-1mm}
    \end{equation}
\end{itemize}
$\mathbf{D}_k(\{f^{\x_c}_k, gf_k\})$ predicts the target value at coordinate $\x_c$ for modality, $k$. 
\vspace{-2mm}
\subsubsection{Self \& Cross Modality Prediction}
\label{subsec:self-cross-pred}
\vspace{-2mm}
% \textcolor{red}{ The math is messy in this part. Please use equations to present the idea, for instance: $\text{input} \to \text{output}$}
Let $\mathbf{X}_1^H \in \mathcal{M}_1^H$ be a data instance with high resolution in modality $\mathcal{M}_1$. With the self-encoder $\mathbf{E}_1^1$ and cross-modal encoder $\mathbf{E}_1^2$ we can get the low dimensional representation of this data instance in both modalities, and consequently achieve continuous super-resolution with the GEI-LIIF based decoder in both modalities, $(\mathbf{FE}_1, \mathbf{G}_1, \mathbf{D}_1), (\mathbf{FE}_2, \mathbf{G}_2, \mathbf{D}_2)$. For example, for a co-ordinate point $\mathbf{x}_c \in \mathbf{X}_c$, 
% \vspace{-2mm}
\begin{equation}
    \mathbf{X}_1^H \to \mathbf{D}_1(\mathbf{FE}_1(\mathbf{E}_1^1(\mathbf{X}_1^H)), \mathbf{G}_1(\mathbf{E}_1^1(\mathbf{X}_1^H)), \mathbf{x}_c)
\end{equation}
% % \vspace{-0.5mm}
represents the prediction at modality $\mathcal{M}_1$ or self-prediction and 
% \vspace{-2mm}
\begin{equation}
    \mathbf{X}_1^H \to \mathbf{D}_2(\mathbf{FE}_2(\mathbf{E}_1^2(\mathbf{X}_1^H)), \mathbf{G}_2(\mathbf{E}_1^2(\mathbf{X}_1^H)), \mathbf{x}_c)
\end{equation} 
% % \vspace{-0.5mm}
represents the prediction at modality $\mathcal{M}_2$ or cross-prediction. 
Similarly, for a data instance $\mathbf{X}_2^H \in \mathcal{M}_2^H$ and for a co-ordinate point $\mathbf{x}_c \in \mathbf{X}_c$, 
% \vspace{-2mm}
\begin{equation}
    \mathbf{X}_2^H \to \mathbf{D}_2(\mathbf{FE}_2(\mathbf{E}_2^2(\mathbf{X}_2^H)), \mathbf{G}_2(\mathbf{E}_2^2(\mathbf{X}_2^H)), \x_c)
\end{equation} 
% % \vspace{-0.5mm}
represents the prediction at modality $\mathcal{M}_2$ or self-prediction and 
% \vspace{-2mm}
\begin{equation}
    \mathbf{X}_2^H \to \mathbf{D}_1(\mathbf{FE}_1(\mathbf{E}_2^1(\mathbf{X}_2^H)), \mathbf{G}_1(\mathbf{E}_2^1(\mathbf{X}_2^H)), \x_c)
\end{equation} 
% % \vspace{-0.5mm}
represents the prediction at modality $\mathcal{M}_1$ or cross-prediction.
\vspace{-4mm}
% =============== %
\section{Results}
\label{sec:results}
\vspace{-2mm}
\subsection{Experimental Setup}
\label{sec:experimental-setup}
\vspace{-2mm}
 \textbullet $\,$ \textbf{Data} The data considered in this paper is generated from the National Renewable Energy Laboratory's Wind Integration National Database (WIND) Toolkit. Specifically, we built the data set for multi-modal super-resolution tasks using simulated wind data. We randomly sampled $1500$ data points from different timestamps among the total available $61368$ instances for each height above from the ground~($10m$, $60m$, $160m$ and $200m$), with $1200$ data points for training the models and $300$ data points for testing. We took wind data from two different heights from the pool of $10m, 60m, 160m$ and $200m$ as the two different modalities. We took two combinations: ($\mathcal{M}_1=10m, \mathcal{M}_2=160m$) and ($\mathcal{M}_1=60m, \mathcal{M}_2=200m$) for doing the experiments. We used bicubic interpolation to generate a pair of high-resolution and super-high-resolution samples for each instance. For example, if the input dimension at both modalities is $(120 \times 160)$, and the super-resolution scale is $1.5 \times$, then the output super-high-resolution dimension is $(180 \times 240)$. \newline
\textbullet $\,$ \textbf{Training} The loss function, $\mathcal{L}$ combines two reconstruction terms,$\mathcal{L}_{self} \text{ and } \mathcal{L}_{cross}$ with a latent term, $\mathcal{L}_{latent}$
\begin{equation}
\label{eq:loss}
\begin{split}
    \mathcal{L} = \mathcal{L}_{self} + \mathcal{L}_{cross} + \mathcal{L}_{latent} 
\end{split}
\end{equation}
The reconstruction terms enable the model to capture signals through both self-prediction and cross-prediction. The latent term promotes the learning of compact, low-dimensional representations.
\begin{equation}
\label{eq:self-loss}
\begin{split}
    &\mathcal{L}_{self} = MSE(\mathbf{D}_1(\mathbf{FE}_1(\mathbf{E}_1^1(\mathbf{X}_1^H))), \mathbf{X}_1^{S}) \\ & + MSE(\mathbf{D}_2(\mathbf{FE}_2(\mathbf{E}_2^2(\mathbf{X}_2^H))), \mathbf{X}_2^{S}) % \vspace{-1mm}
\end{split}
\end{equation}
\begin{equation}
\label{eq:cross-loss}
\begin{split}
    &\mathcal{L}_{cross} = MSE(\mathbf{D}_1(\mathbf{FE}_1(\mathbf{E}_1^2(\mathbf{X}_2^H))), \mathbf{X}_1^{S}) \\ & + MSE(\mathbf{D}_2(\mathbf{FE}_2(\mathbf{E}_1^2(\mathbf{X}_1^H))), \mathbf{X}_2^{S}) % \vspace{-1mm}
\end{split}
\end{equation}
\begin{equation}
\label{eq:latent-loss}
\begin{split}
    &\mathcal{L}_{latent} = MSE\left(\mathbf{E}_1^1(\mathbf{X}_1^H), \dfrac{\mathbf{E}_1^1(\mathbf{X}_1^H) + \mathbf{E}_2^1(\mathbf{X}_2^H)}{2}\right) \\
    & + MSE\left(\mathbf{E}_2^1(\mathbf{X}_2^H), \dfrac{\mathbf{E}_1^1(\mathbf{X}_1^H) + \mathbf{E}_2^1(\mathbf{X}_2^H)}{2}\right) \\
    & + MSE\left(\mathbf{E}_2^2(\mathbf{X}_2^H), \dfrac{\mathbf{E}_2^2(\mathbf{X}_2^H) + \mathbf{E}_1^2(\mathbf{X}_1^H)}{2}\right) \\
    & + MSE\left(\mathbf{E}_1^2(\mathbf{X}_1^H), \dfrac{\mathbf{E}_2^2(\mathbf{X}_2^H) + \mathbf{E}_1^2(\mathbf{X}_1^H)}{2}\right) 
\end{split}
\end{equation}
Unlike CLUE~\cite{clue}, we do not enforce our model to learn a unified latent space representation which gives us the freedom to bypass the need for a modality classifier model and an adversarial loss function for optimization.

\vspace{-5mm}
\subsection{Observed Results}
\label{subsubsec:observed-results}
\vspace{-2.5mm}
We tested the performance of our model at different super-resolution scales for both self and cross predictions on the test dataset consisting $300$ datapoints. The high resolution input dimension was set to $(1 \times 120 \times 160)$ and the low resolution representation had a dimension of $(1 \times 15 \times 20)$. We used a pretrained ResNet18 \cite{resnet18} model as the global encoder that encodes the low dimensional representation and extracts the features as a vector, and finetuned the weights while optimizing the other parts of the model. 
% Table \ref{table:proj-result} shows the results of our proposed methodology with labels GEI-LIIF. Figure \ref{fig:result-plot-60-200} shows one specific case of cross prediction results at finer super resolution scales. Figure \ref{fig:results} show some visual results of the predicted super resolution samples at different superresolution scales with $\mathcal{M}_{in}=10m$ and $\mathcal{M}_{out}=160m$ for northern projection of wind data.
\vspace{-5mm}
\subsubsection{Super-Resolution Performance}
\label{subsec:super-resolution-result}
\begin{figure}[t]
\begin{center}
    \includegraphics[width=0.425\textwidth]{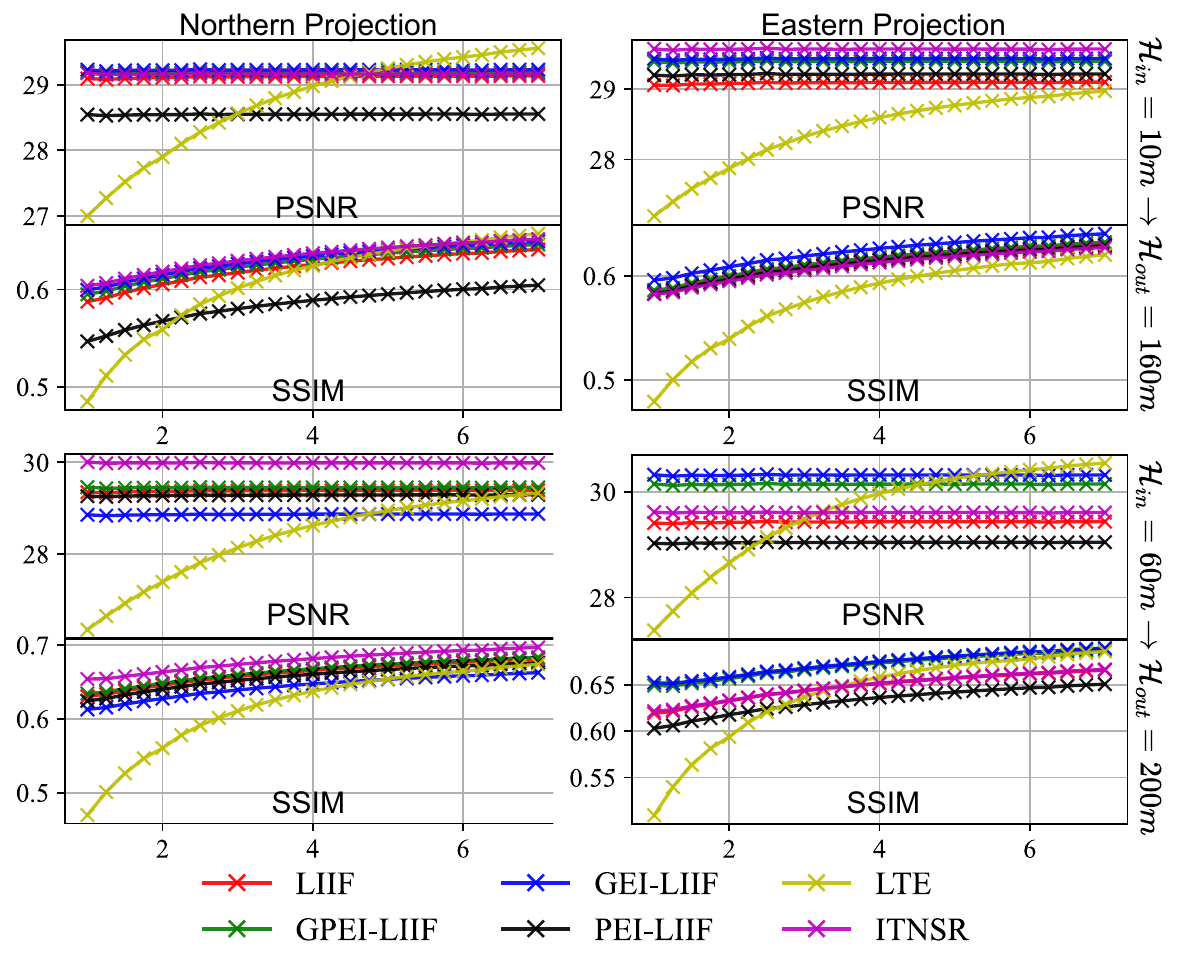}
    \vspace{-4.5mm}
    \caption{Super-resolution performance of various decoders.}
    \label{fig:result}
\end{center}
\vspace{-9mm}
\end{figure}

\vspace{-2.5mm}
We tested the super-resolution performance of our designed GEI-LIIF decoder by replacing it with various other super-resolution models. We also made some modifications in our proposed methodology and compared the performances with these modified models. As the baseline, we chose LIIF based decoder for modality specific super-resolution where the decoder only takes the extracted local features as its input. Positional encoders are functions of the 2-D coordinates based on Fourier based positional encoding. $\mathbf{P}:\X_c \in \mathbb{R}^2 \to \mathcal{PF} \in \mathbb{R}^p$ is the positional encoder with $\x_c \in \X_c$ as its input with $p$ as the dimension of the positional encoder output. We design PEI-LIIF~(Positional Encoding Incorporated Local Implicit Image Function) decoder, $\mathbf{D}_k(\{f^{\x_c}_k, \mathbf{P}(\x_c)\})$ and GPEI-LIIF~(Global \& Positional Encoding Incorporated Local Implicit Image Function), $\mathbf{D}_k(\{f^{\x_c}_k, gf_k, \mathbf{P}(\x_c)\})$. PEI-LIIF uses extracted local features and positional encodings as its input, whereas GPEI-LIIF uses extracted local features, global features and positional encodings as its input. We also compared our designed decoder with local texture estimator based decoder (LTE) \cite{lte}, and implicit transformer network based decoder (ITNSR) \cite{itnsr}. Figure \ref{fig:result} shows the comparative results for the cross-prediction scenarios where the input modality height is closer to the ground and the output modality height is much higher above from the ground. 
% % \vspace{-1mm}

% \begin{figure}[t]
% \begin{center}
%     \includegraphics[width=0.45\textwidth]{figs/SR.pdf}
%     % % \vspace{-9mm}
%     \caption{Super-resolution performances for various decoders.}
%     \label{fig:result}
% \end{center}
% % % \vspace{-10mm}
% \end{figure}

% \begin{figure}[t]
% \begin{center}
%     \includegraphics[width=\textwidth]{figs/SR-result.pdf}
%     % % \vspace{-9mm}
%     \caption{Super-resolution performances for various decoders with northern projection.}
%     \label{fig:result-ua}
% \end{center}
% % % \vspace{-10mm}
% \end{figure}

% \begin{figure}[t]
% \begin{center}
%     \includegraphics[width=0.45\textwidth]{figs/va/VA.pdf}
%     % % \vspace{-9mm}
%     \caption{Super-resolution performances for various decoders with eastern projection.}
%     \label{fig:result-va}
% \end{center}
% % % \vspace{-10mm}
% \end{figure}

% =============== %
\vspace{-5mm}
\subsubsection{Compression Performance}
\label{subsec:compression-result}
\vspace{-2.5mm}
\begin{figure}[t]
\begin{center}
    \includegraphics[width=0.385\textwidth]{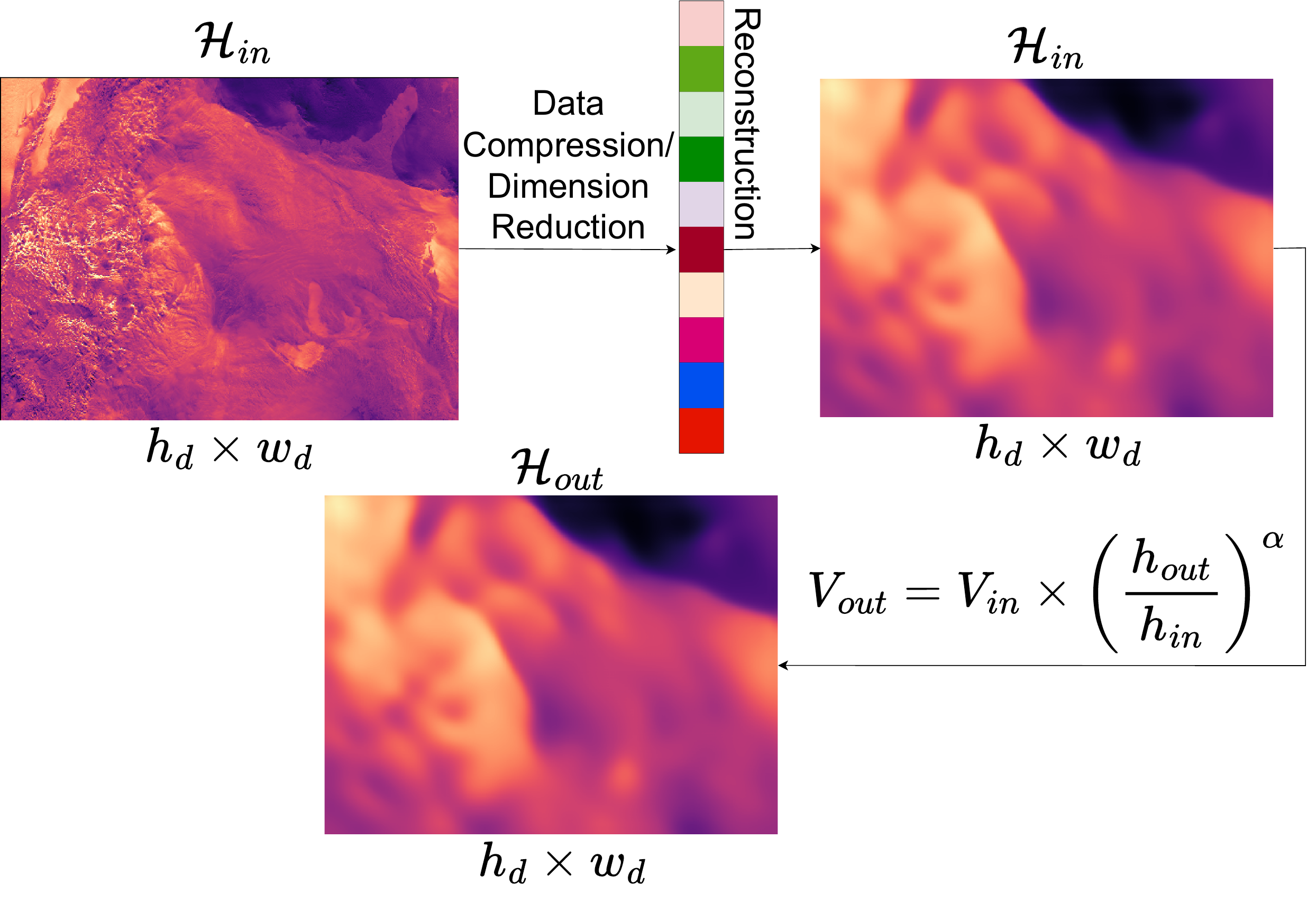}
    \vspace{-5mm}
    \caption{Data compression followed by cross-modal (i.e., cross-altitude) prediction using wind power law.}
    \label{fig:method-compression}
\end{center}
\vspace{-10mm}
\end{figure}
We tested our approach and compared its compression performance with other data compression methods. We used prediction by the partial matching (PPM) data compression algorithm with the $\mu$-law based encoding at different quantization levels ($Q$) to compress and reconstruct data \cite{ppm}. We also tested bicubic interpolation to compress and decompress the data. For comparison of cross-modal predictions, we used the wind power law to transform the reconstructed data at one height to another height according to the equation, $\frac{v_1}{v_2} = \left(\frac{h_1}{h_2}\right)^\alpha$ \cite{wind-power-law}, with the PPM and bicubic methods. Figure \ref{fig:method-compression} summarizes the cross-modal prediction performances by these methods. Table \ref{tab:compression-result} shows the data compression performance for different methods. We set $\mu=255$ and $\alpha=0.16$ for the PPM and bicubic methods. As this workflow is only capable of reconstructing the output as the same dimension of the input, we set the super-resolution scale $s=1$ for GEI-LIIF and GPEI-LIIF models for a fair comparison. Unlike super-resolution results, we also tested dimension reduction factor $d=4$ for bicubic, GEI-LIIF and GPEI-LIIF models to see how these methods perform compared to compression methods when dimension reduction factor is smaller.

\begin{table}[t]
    \centering
    \begin{tabular}{|c|c|c|c|}
        \hline
        \multicolumn{4}{|c|}{Northern Projection} \\ %& \multicolumn{4}{|c|}{Eastern Projection} \\
        \hline
        Method & CR $\uparrow$ & PSNR $\uparrow$ & SSIM $\uparrow$ \\ %& Method & CR $\uparrow$ & PSNR $\uparrow$ & SSIM $\uparrow$\\
        \hline 
        \multicolumn{4} {|c|} {$\mathcal{H}_{in} = 10m \rightarrow \mathcal{H}_{out}=160m$} \\
        
        \hline
        $\text{PPM}_{Q=8}$ & 95.0584 & 24.1419 & 0.4104 \\
        \hline
        $\text{PPM}_{Q=16}$ & 92.8557 & 28.3741 & 0.6147 \\
        \hline

        $\text{Bicubic}_{d=8}$ & 98.4375 & 28.6436 & 0.5506 \\
        \hline
        $\text{Bicubic}_{d=4}$ & 93.7500 & 29.4141 & 0.6157 \\
        \hline
        $\text{GEI-LIIF}_{d=8}$ & \textcolor{blue}{98.4375} & 29.2306 & 0.5996 \\
        \hline
        % $\text{Bicubic}_{d=4}$ & 93.7500 & 29.4141 & 0.6157 \\
        % \hline
        $\text{GPEI-LIIF}_{d=4}$ & 93.7500 & \textcolor{blue}{29.6598} & \textcolor{blue}{0.6407} \\
        \hline

        % \multicolumn{8} {|c|} {$\mathcal{H}_{in} = 10m \rightarrow \mathcal{H}_{out}=200m$} \\
        % \hline
        % $\text{PPM}_{Q=8}$ & 95.0610 & 23.6673 & 0.3893 & $\text{PPM}_{Q=8}$ & 94.8042 & 23.4366 & 0.4275\\
        % \hline
        % $\text{PPM}_{Q=16}$ & 92.8751 & 27.5924 & 0.5755 & $\text{PPM}_{Q=16}$ & 92.5828 & 27.6692 & 0.6114\\
        % \hline

        % $\text{Bicubic}_{d=8}$ & 98.4375 & 28.0937 & 0.5341 & $\text{Bicubic}_{d=8}$ & 98.4375 & 27.9385 & 0.5185\\
        % \hline
        % $\text{GPEI-LIIF}_{d=8}$ & 98.4375 & 28.8448 & 0.5963 & $\text{GPEI-LIIF}_{d=8}$ & 98.4375 & 29.4261 & 0.5935\\
        % \hline 
        % $\text{Bicubic}_{d=4}$ & 93.7500 & 28.7553 & 0.5914 & $\text{Bicubic}_{d=4}$ & 93.7500 & 28.9412 & 0.5918\\
        % \hline
        % $\text{GPEI-LIIF}_{d=4}$ & 93.7500 & 29.1662 & 0.6383 & $\text{GPEI-LIIF}_{d=4}$ & 93.7500 & 29.9830 & 0.6415\\
        % \hline 

        % \multicolumn{8} {|c|} {$\mathcal{H}_{in} = 60m \rightarrow \mathcal{H}_{out}=160m$} \\
        % \hline
        % $\text{PPM}_{Q=8}$ & 95.7095 & 24.6593 & 0.4904 & $\text{PPM}_{Q=8}$ & 95.3401 & 23.8341 & 0.5210\\
        % \hline
        % $\text{PPM}_{Q=16}$ & 93.5867 & 30.0943 & 0.7205 & $\text{PPM}_{Q=16}$ & 93.2037 & 30.1731 & 0.7512\\
        % \hline

        % $\text{Bicubic}_{d=8}$ & 98.4375 & 30.3703 & 0.6299 & $\text{Bicubic}_{d=8}$ & 98.4375 & 29.8631 & 0.6083\\
        % \hline
        % $\text{GEI-LIIF}_{d=8}$ & 98.4375 & 30.3876 & 0.6591 & $\text{GEI-LIIF}_{d=8}$ & 98.4375 & 29.9155 & 0.6431\\
        % \hline
        % $\text{Bicubic}_{d=4}$ & 93.7500 & 31.7323 & 0.7168 & $\text{Bicubic}_{d=4}$ & 93.7500 & 31.4453 & 0.7088\\
        % \hline
        % $\text{GPEI-LIIF}_{d=4}$ & 93.7500 & 30.8012 & 0.7210 & $\text{GEI-LIIF}_{d=4}$ & 93.7500 & 31.4341 & 0.7316\\
        % \hline 

        \multicolumn{4} {|c|} {$\mathcal{H}_{in} = 60m \rightarrow \mathcal{H}_{out}=200m$} \\
        \hline
        $\text{PPM}_{Q=8}$ & 95.7176 & 24.2708 & 0.4699 \\
        \hline
        $\text{PPM}_{Q=16}$ & 93.6066 & 29.2434 & 0.6793 \\
        \hline

        $\text{Bicubic}_{d=8}$ & 98.4375 & 29.8462 & 0.6144 \\
        \hline
        $\text{Bicubic}_{d=4}$ & 93.7500 & \textcolor{blue}{31.0304} & 0.6917 \\
        \hline
        $\text{GPEI-LIIF}_{d=8}$ & \textcolor{blue}{98.4375} & 29.4583 & 0.6342 \\
        \hline 
        % $\text{Bicubic}_{d=4}$ & 93.7500 & 31.0304 & 0.6917 \\
        % \hline
        $\text{GPEI-LIIF}_{d=4}$ & 93.7500 & 30.7283 & \textcolor{blue}{0.7049} \\
        \hline 
%     \end{tabular}
    % \vspace{-2.5mm}
%     \caption{Comparative analysis of data compression performance of different schemes for northern projection.}
    % \vspace{-5mm}
%     \label{tab:compression-result-ua}
% \end{table}

% \begin{table}[t]
%     \centering
%     \begin{tabular}{|c|c|c|c|c|c|c|c|}
        % \hline
        \multicolumn{4}{|c|}{Eastern Projection} \\ %& \multicolumn{4}{|c|}{Eastern Projection} \\
        \hline
        Method & CR $\uparrow$ & PSNR $\uparrow$ & SSIM $\uparrow$ \\ %& Method & CR $\uparrow$ & PSNR $\uparrow$ & SSIM $\uparrow$\\
        \hline 
        \multicolumn{4} {|c|} {$\mathcal{H}_{in} = 10m \rightarrow \mathcal{H}_{out}=160m$} \\
        
        \hline
        $\text{PPM}_{Q=8}$ & 94.8006 & 23.8402 & 0.4493\\
        \hline
        $\text{PPM}_{Q=16}$ & 92.5589 & 28.4303 & 0.6527\\
        \hline

        $\text{Bicubic}_{d=8}$ & 98.4375 & 28.4412 & 0.5379\\
        \hline
        $\text{Bicubic}_{d=4}$ & 93.7500 & 29.5427 & 0.6191\\
        \hline
        $\text{GEI-LIIF}_{d=8}$ & \textcolor{blue}{98.4375} & 29.433 & 0.5959\\
        \hline
        % $\text{Bicubic}_{d=4}$ & 93.7500 & 29.5427 & 0.6191\\
        % \hline
        $\text{GEI-LIIF}_{d=4}$ & 93.7500 & \textcolor{blue}{30.4430} & \textcolor{blue}{0.6646}\\
        \hline

        % \multicolumn{8} {|c|} {$\mathcal{H}_{in} = 10m \rightarrow \mathcal{H}_{out}=200m$} \\
        % \hline
        % $\text{PPM}_{Q=8}$ & 95.0610 & 23.6673 & 0.3893 & $\text{PPM}_{Q=8}$ & 94.8042 & 23.4366 & 0.4275\\
        % \hline
        % $\text{PPM}_{Q=16}$ & 92.8751 & 27.5924 & 0.5755 & $\text{PPM}_{Q=16}$ & 92.5828 & 27.6692 & 0.6114\\
        % \hline

        % $\text{Bicubic}_{d=8}$ & 98.4375 & 28.0937 & 0.5341 & $\text{Bicubic}_{d=8}$ & 98.4375 & 27.9385 & 0.5185\\
        % \hline
        % $\text{GPEI-LIIF}_{d=8}$ & 98.4375 & 28.8448 & 0.5963 & $\text{GPEI-LIIF}_{d=8}$ & 98.4375 & 29.4261 & 0.5935\\
        % \hline 
        % $\text{Bicubic}_{d=4}$ & 93.7500 & 28.7553 & 0.5914 & $\text{Bicubic}_{d=4}$ & 93.7500 & 28.9412 & 0.5918\\
        % \hline
        % $\text{GPEI-LIIF}_{d=4}$ & 93.7500 & 29.1662 & 0.6383 & $\text{GPEI-LIIF}_{d=4}$ & 93.7500 & 29.9830 & 0.6415\\
        % \hline 

        % \multicolumn{8} {|c|} {$\mathcal{H}_{in} = 60m \rightarrow \mathcal{H}_{out}=160m$} \\
        % \hline
        % $\text{PPM}_{Q=8}$ & 95.7095 & 24.6593 & 0.4904 & $\text{PPM}_{Q=8}$ & 95.3401 & 23.8341 & 0.5210\\
        % \hline
        % $\text{PPM}_{Q=16}$ & 93.5867 & 30.0943 & 0.7205 & $\text{PPM}_{Q=16}$ & 93.2037 & 30.1731 & 0.7512\\
        % \hline

        % $\text{Bicubic}_{d=8}$ & 98.4375 & 30.3703 & 0.6299 & $\text{Bicubic}_{d=8}$ & 98.4375 & 29.8631 & 0.6083\\
        % \hline
        % $\text{GEI-LIIF}_{d=8}$ & 98.4375 & 30.3876 & 0.6591 & $\text{GEI-LIIF}_{d=8}$ & 98.4375 & 29.9155 & 0.6431\\
        % \hline
        % $\text{Bicubic}_{d=4}$ & 93.7500 & 31.7323 & 0.7168 & $\text{Bicubic}_{d=4}$ & 93.7500 & 31.4453 & 0.7088\\
        % \hline
        % $\text{GPEI-LIIF}_{d=4}$ & 93.7500 & 30.8012 & 0.7210 & $\text{GEI-LIIF}_{d=4}$ & 93.7500 & 31.4341 & 0.7316\\
        % \hline 

        \multicolumn{4} {|c|} {$\mathcal{H}_{in} = 60m \rightarrow \mathcal{H}_{out}=200m$} \\
        \hline
        $\text{PPM}_{Q=8}$ & 95.3462 & 23.6018 & 0.5028\\
        \hline
        $\text{PPM}_{Q=16}$ & 93.2222 & 29.3593 & 0.7089\\
        \hline

        $\text{Bicubic}_{d=8}$ & 98.4375 & 29.3787 & 0.5887\\
        \hline
        $\text{Bicubic}_{d=4}$ & 93.7500 & \textcolor{blue}{30.8214} & 0.6800\\
        \hline
        $\text{GEI-LIIF}_{d=8}$ & \textcolor{blue}{98.4375} & 30.3195 & 0.6522\\
        \hline 
        % $\text{Bicubic}_{d=4}$ & 93.7500 & 30.8214 & 0.6800\\
        % \hline
        $\text{GPEI-LIIF}_{d=4}$ & 93.7500 & 30.8097 & \textcolor{blue}{0.6818}\\
        \hline 
    \end{tabular}
    \vspace{-2mm}
    \caption{Comparative analysis of data compression performance of different schemes.}
    \vspace{-5.25mm}
    \label{tab:compression-result}
\end{table}

\vspace{-5.2mm}
\subsection{Discussion}
\label{subsec:discussion}
\vspace{-2.5mm}
Figure \ref{fig:result} shows the super-resolution performances for different decoders for 8 different cases (2 projections, 2 different cross-modal scenarios for each projection, 2 metrics for each cross-modal scenario). Among them, GEI-LIIF comes out to be the best performing one in 4 cases, whereas ITNSR decoder does the best in the other 4 cases. At some extreme scales, LTE beats other models but performs poorly in other super-resolution scales. In terms of compression performance, our proposed model (either with GEI-LIIF or GPEI-LIIF decoder) outperforms PPM or Bicubic models in terms of compression ratio. The compression models achieve high PSNR and SSIM only when the compression ratios are the lowest. These models are not capable of achieving the best performance in all three metrics simultaneously. 
Due to space limitations, we do not show the model performance in other cross-modal scenarios where the input height modality is much higher above the ground and the output modality is much closer to the ground as those cases are not of greater concern compared to its counterpart, nor the self prediction cases. But the results in those cases are similar to what we see in this cross-modal scenario, both in terms of super-resolution and compression performance. 
In terms of super-resolution performance, GEI-LIIF is not always beating its counterparts, rather ITNSR comes out to be the champion in the same number of cases. This indicates that there should be a better way of fusing the global and local features at the decoding stage to achieve better super-resolution performance, and it still remains an open question.
\vspace{-5.2mm}
\section{Conclusion}
\label{sec:conclusion}
\vspace{-4mm}
We proposed a novel deep learning solution for simultaneous continuous super-resolution, data dimensionality reduction, and multi-modal learning of climatological data. We specifically developed a local implicit neural network model for learning continuous, rather than discrete, representations of climate data, such as wind velocity fields used for wind farm power modeling across the continental United States, along with multi-modal dimension reducing encoder that facilitates dimension reduction and cross modality extrapolation. We also introduced a latent loss function to ensure cross modality learning. Obtained results have shown the promising potential to solve real-world scenarios in wind energy resource assessment for electricity generation, efficient storage of huge amount of data by dimensionality reduction, and extrapolation of data to inaccessible spatial spaces (e.g., specific altitudes) from available wind data. However, our model is feasible only for a small number of modalities as the total number of encoders will increase quadratically with the increase of modalities. Designing a more scalable model that can handle a higher number of modalities is a topic for future research.
\vspace{-2mm}
\bibliographystyle{IEEEbib}
\vspace{-3mm}
\bibliography{strings,refs}
\vspace{-2mm}
\end{document}